\documentclass[sigconf]{acmart}
\AtBeginDocument{%
  }

\copyrightyear{2025}
\acmYear{2025}
\setcopyright{acmlicensed}
\acmConference[McGE '25] {Proceedings of the 3rd International Workshop on Multimedia Content Generation and Evaluation: New Methods and Practice}{October 27--31, 2025}{Dublin, Ireland.}
\acmBooktitle{Proceedings of the 3rd International Workshop on Multimedia Content Generation and Evaluation: New Methods and Practice (McGE '25), October 27--31, 2025, Dublin, Ireland}
\acmISBN{979-8-4007-2060-4/2025/10}
\acmDOI{10.1145/3746278.3759387}

\acmSubmissionID{mcge22}

\usepackage{subcaption}
\usepackage{amsmath}
\usepackage{amsfonts}
\usepackage{xcolor} 
\usepackage{colortbl} 
\usepackage{color,soul}
\usepackage{tikz}
\usepackage{lipsum}
\usetikzlibrary{arrows.meta, positioning}
\usepackage{balance}

\begin{document}

\title{When and How to Cut Classical Concerts? \\ A Multimodal Automated Video Editing Approach}

\author{Daniel Gonzálbez-Biosca}
\email{danielgonzalbez@uoc.edu}
\affiliation{%
  \institution{eHealth Center, Faculty of Computer Science, Multimedia and Telecommunications, Universitat Oberta de Catalunya}
  \streetaddress{Rambla del Poblenou, 156}
  \city{Barcelona}
  \state{Catalonia}
  \postcode{08018}
  \country{Spain}
}

\author{Josep Cabacas-Maso}
\email{jcabacas@uoc.edu}
\affiliation{%
  \institution{eHealth Center, Faculty of Computer Science, Multimedia and Telecommunications, Universitat Oberta de Catalunya}
  \streetaddress{Rambla del Poblenou, 156}
  \city{Barcelona}
  \state{Catalonia}
  \postcode{08018}
  \country{Spain}
}

\author{Carles Ventura}
\email{cventuraroy@uoc.edu}
\orcid{0000-0002-3055-339X}
\affiliation{%
  \institution{eHealth Center, Faculty of Computer Science, Multimedia and Telecommunications, Universitat Oberta de Catalunya}
  \streetaddress{Rambla del Poblenou, 156}
  \city{Barcelona}
  \state{Catalonia}
  \postcode{08018}
  \country{Spain}
}

\author{Ismael Benito-Altamirano}
\email{ibenitoal@uoc.edu}
\orcid{0000-0002-2504-6123}
\affiliation{%
  \institution{eHealth Center, Faculty of Computer Science, Multimedia and Telecommunications, Universitat Oberta de Catalunya}
  \streetaddress{Rambla del Poblenou, 156}
  \city{Barcelona}
  \state{Catalonia}
  \postcode{08018}
  \country{Spain}
}
\affiliation{%
  \institution{MIND/IN2UB, Department of Electronic and Biomedical Engineering, Universitat de Barcelona}
  \streetaddress{Carrer de Martí i Franquès, 1}
  \city{Barcelona}
  \state{Catalonia}
  \postcode{08028}
  \country{Spain}
}

\renewcommand{\shortauthors}{Gonzálbez-Biosca et al.}

\begin{abstract}
    Automated video editing remains an underexplored task in the computer vision and multimedia domains, especially when contrasted with the growing interest in video generation and scene understanding. In this work, we address the specific challenge of editing multicamera recordings of classical music concerts by decomposing the problem into two key sub-tasks: \emph{when to cut} and \emph{how to cut}. Building on recent literature, we propose a novel multimodal architecture for the temporal segmentation task (\emph{when to cut}), which integrates log-mel spectrograms from the audio signals, plus an optional image embedding, and scalar temporal features through a lightweight convolutional-transformer pipeline. For the spatial selection task (\emph{how to cut}), we improve the literature by updating from old backbones, e.g. ResNet, with a CLIP-based encoder and constraining distractor selection to segments from the same concert. Our dataset was constructed following a pseudo-labeling approach, in which raw video data was automatically clustered into coherent shot segments. We show that our models outperformed previous baselines in detecting cut points and provide competitive visual shot selection, advancing the state of the art in multimodal automated video editing.
\end{abstract}





\maketitle

\section{Introduction}\label{sec:introduction}

Automated video editing is an open challenge in computer vision and multimedia fields. Within it, authors have approached this problem from different perspectives, such as: scene detection and segmentation~\cite{del2013state,zhou2022survey} or image or video inpainting~\cite{zhang2024minutes,quan2024deep}, which, indeed, are related to the underlying idea of understanding and tempering with multicamera videos, but they fall behind the actual idea of video editing a whole video production from its sources. In fact, it is surprising that modern approches have arisen to directly create videos from scratch using LLM architectures, by using prompts to guide the generation~\cite{henschel2025streamingt2v}, but at some point, researchers left behind the actual problem of video editing. As most of the literuture focus on generating new videos, or the understanding of existing ones, but not on the actual task of editing a set of video segments into a final video.

\begin{figure}[ht]
    \centering
    \includegraphics[width=0.20\textwidth]{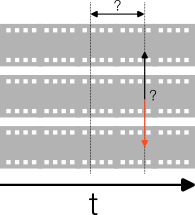}
    \caption{Overview of the editing task. Given a multicamera recording of a classical music concert, the system must decide \emph{when to cut} (horizontal arrow) and \emph{how to cut} (vertical arrow): that is, which shot transition to perform, and which view to show next.}
    \label{fig:editing_problem}
\end{figure}

In 2022, Jimenez et al.~\cite{jimenez2022automated}, working towards video editing, proposed an idea of splitting the automated video editing problem into to halves: first, a problem of ``\emph{when to cut?}''; and second, a problem on ``\emph{how to cut?}'' (see~\autoref{fig:editing_problem}). For the first problem, they introduced an statical approach of measuring a distribution from their actual data and just sampling from that distribution as a first approach to that task; and, for the later, they devised a method to use different CNN-based back-bone models plus different attention mechanisms to select the best cut selection to jump from a video segment to another. More recently, in 2024, Lin et al.~\cite{lin2024videogenic} presented a solution to detect the best moments to cut a video, using CLIP, a well-known multimodal model that combines visual and textual information, but they focused more in the content of the shots and understanding the capabilities of CLIP as a model to detect such type of contents. Related to this research path, in 2025, Caravaca et al.~\cite{caravaca2025vclipper} presented a similar work tackling the multimodal component of CLIP, similarly to Lin et al.~\cite{lin2024videogenic}, while using also the text embeddings of CLIP in their architecture. Recently, Lee et al.~\cite{lee2024pseudo} presented a similar solution to Jimenez et al.~\cite{jimenez2022automated}, where they focused on the problem of generating a dataset from edited videos into a pseudo-annotated dataset for the task we want to tackle, automated video editing.

In this work, we propose a multimodal approach to automated video editing of classical music concerts that builds upon and extends recent state-of-the-art methods. Our dataset was created in a similar fashion to Jimenez et al.~\cite{jimenez2022automated}. Thus, we downloaded a series of videos from the internet and processed them in a similar fashion to Lee et al.~\cite{lee2024pseudo} by pseudo-labeling the videos by clustering the segments of videos into different clusters, as in representations of the same shot.

To determine \emph{when to cut} a video, we introduced our own architecture based on transformers that jointly processes the audio input windowed and transformed into log-mel spectrogram and a scalar embedding of the time elapsed since the last cut. Preprocessing audio into spectrograms is a well-known technique. For example, there exist applications like DeepSpectrum, originally applied to snores~\cite{amiriparian2017snore}, that has been applied to detect emotion in Spanish speakers~\cite{ortega2024better}. Then the spectrogram was fed through a series of one-dimensional convolutional blocks followed by transformer layers, capturing both temporal and frequential patterns in the input. Simultaneously, a scalar input representing the time passed since the last cut has presented is projected into a learned embedding space and enhanced with an additive identifier vector. The resulting audio and scalar embeddings were concatenated and fed into a feed-forward network that outputs a cut probability. We also propose a multimodal variant of this approach which also adds information from the visual part of the video by using a CLIP-like encoder to extract visual features from the video frames.

To address \emph{how to cut} a video, we reproduced and refined the method proposed by Jimenez et al.~\cite{jimenez2022automated}, replacing their ResNet-based backbone with a CLIP-based encoder to improve the alignment between visual features and high-level semantics. We differ on how Jiminez et al.~\cite{jimenez2022automated} trained their models by using the pseudo-labeled technique from Lee et al.~\cite{lee2024pseudo}. This implies clustering the pseudo-scenes and building the test dataset using videos only from the same concert, which is a more realistic scenario than using random videos from the entire dataset.


All in all, our architecture tackles the problem of automated video editing by combining visual and audio features from a pseudo-labeled dataset of classical music concerts, reproduces the SOTA and/or baseline results and even improves them for the task of when to cut a video segment. For the \emph{when to cut} task our models outperformed the proposed baselines with more than 10 percentual points, best result in the test dataset for this task was obtained in with the unimodal model achieving an accuracy of 62.01\% over the 49.42\% of the Poisson baseline. Regarding the \emph{how to cut} task, our model achieved a Recall@1 of 28.49\% over the 11.18\% of the ResNet baseline, and a Recall@3 of 51.97\% over the 30.08 of the ResNet baseline, slightly pairing but outperforming the Xception baseline.

\section{Methodology}\label{sec:methodology}
\subsection{Task definition}\label{subsec:task-definition}

Automated video editing is a complex task that can be decomposed into several sub-tasks. In this work, we focus on two key sub-tasks: \emph{when to cut} and \emph{how to cut}. On the one hand, deciding \emph{when to cut} a video segment involves identifying the most appropriate moments to transition between different shots or scenes. This task can be seen as a \emph{temporal segmentation problem}, this is the inverse problem to scene segmentation~\cite{zhou2022survey}, where the goal is to identify the boundaries between different segments of a video. The task can be handled by determining the most iconic moments in a video, such as a ``videogenic'' metric~\cite{lin2024videogenic}; or by extracting features from the audio component of the videos, i.e. in a concert setting all the multicamera recordings are synchronized, so the audio component can be used to determine the best moments to cut by extracting sound features like the log-mel spectrogram~\cite{mehta2021music}.

On the other hand, deciding \emph{how to cut} a video segment involves selecting the most appropriate shot or view to show next. This task can be seen as a \emph{spatial selection problem}, where the goal is to select the best shot from a set of available shots. In our case, we focus on the spatial selection of shots in multicamera recordings of classical music concerts, where each camera captures a different view of the same scene. The task can be handled by extracting visual features from the video frames, i.e. using CLIP~\cite{lin2024videogenic,caravaca2025vclipper}, and selecting the best shot by a distractor-based selection mechanism, which is a method to select the best shot from a set of candidates based on the visual features extracted from the video frames~\cite{jimenez2022automated}.

\subsection{Dataset download and labelling}\label{subsec:dataset-download-and-labelling}

\begin{figure}[!ht]
    \centering
    \resizebox{\columnwidth}{!}{%
        \begin{tikzpicture}[
            node distance=0.8em and 1.5em, 
            box/.style={
                rectangle, draw, rounded corners=3pt,
                align=center, font=\small,
                text width=0.20\textwidth,
                minimum height=2em
            },
            arrow/.style={-Latex, thick}
        ]

            \node[box] (selection) {Curated List of \\ \textbf{100 Classical Music Videos}};
            \node[box, below=0.5 of selection] (yt) {\texttt{yt-dlp} Downloader (Python) \\ nHD 360p videos (\textbf{640x360})};

            \node[box, below left=1.5em and -3em of yt] (audio_raw) {Audio waveform extracted\\ (44.1/48 kHz)};
            \node[box, below right=1.5em and -3em of yt] (video_raw) {Video frames extracted\\ (24/30 FPS)};

            \node[box, below=1.5em of audio_raw] (resample) {\texttt{ffmpeg} resampling to\\ \textbf{16 kHz}};
            \node[box, below=1.5em of video_raw] (frame_extract) {\texttt{OpenCV} frame extraction at\\ \textbf{5 FPS}};

            \node[box, below=10em of yt, text width=0.30\textwidth] (final_dataset) {\textbf{Unlabeled multimodal dataset}\\ (audio + visual data per video)};

            \draw[arrow] (selection) -- (yt);
            \draw[arrow] (yt) -- (audio_raw);
            \draw[arrow] (audio_raw) -- (resample);
            \draw[arrow] (resample) -- (final_dataset);
            \draw[arrow] (yt) -- (video_raw);
            \draw[arrow] (video_raw) -- (frame_extract);
            \draw[arrow] (frame_extract) -- (final_dataset);

        \end{tikzpicture}
    }
    \caption{Data collection pipeline for the classical concert video dataset. Videos were downloaded in nHD resolution (360p) using the \texttt{yt-dlp} downloader. The audio was resampled to 16 kHz using \texttt{ffmpeg}, and video frames were extracted at 5 FPS using \texttt{OpenCV}. This intermediate dataset consists of unlabeled multimodal data, including audio and visual components for each video.}
    \label{fig:data_collection}
\end{figure}
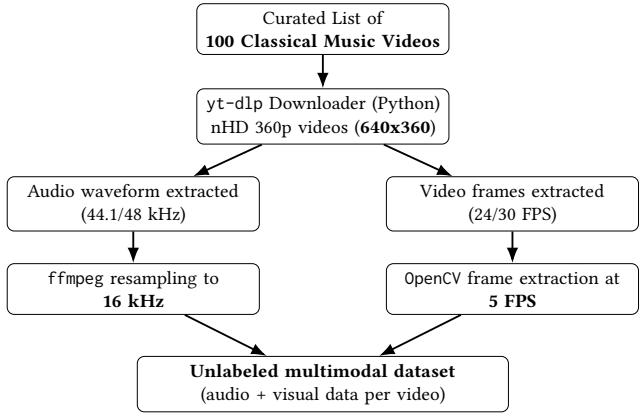

We curated a dataset consisting of 100 online videos of classical music concerts~\footnote{List and downloader avalilable on request}, ensuring that no two recordings corresponded to the same musical piece in order to enhance diversity and avoid redundancy. For this task we implemented a custom downloader that used Python as programming language and the \texttt{yt-dlp}~\cite{yt-dlp} library at its core. This downloader was able to work in different scenarios, e.g downloading whole videos which descriptions contained pointers to cut the video in different songs or playlist that contained a video per each song. Concert duration varied, ranging from 10 to 90 minutes.~\autoref{fig:data_collection} shows the pipeline used to preprocess the multimodal dataset: videos were downloaded in nHD quality, namely this is the 360p 16:9 resolition, e.g. 640x360 pixel resolution, directly using the downloading tool. This ensured the spatial downsampling was performed equally by the remote platform , and not by us, which could have introduced artifacts in the videos. Later, we used \texttt{ffmpeg} to resample the audio tracks to 16 kHz, standardizing the input, as most original recordings had a range of sampling rates up to 44 kHz. This resampling was necessary to ensure that the audio features extracted from the videos were consistent across all recordings. This is a common practice and other authors perform previous to use the audio in a deep learning pipeline~\cite{mensah2025deep}. Finally, we chose to work with 5 FPS snapshots from the original videos, to reduce computational costs and ensure that the model could process the videos in a reasonable time frame.

Regarding the segmentation of the edited videos into scenes for labelling, we implemented a lightweight yet effective Python pipeline that automatically detected shot changes. While libraries such as \texttt{scenedetect}~\cite{scenedetect} are commonly used for this task—relying on color histogram analysis and thresholding to identify abrupt visual transitions—they often lack semantic awareness, focusing instead on low-level pixel variations. We splited on how to label the dataset for the two tasks, \emph{when to cut} and \emph{how to cut}, as follows.

On the one hand, for the~\emph{spatial selection task} (\emph{how to cut}), we designed a frame differencing method that operates by sequentially converting each frame to grayscale and computing the absolute difference between adjacent frames, extracting a mean pixel-wise difference score. A scene change is detected when this score exceeds a predefined threshold, indicating a substantial shift in visual content. To ensure temporal stability and avoid saving frames during abrupt transitions, we instead recorded the frame occurring approximately two seconds before the detected change, with the offset computed using the video's frame rate (FPS).

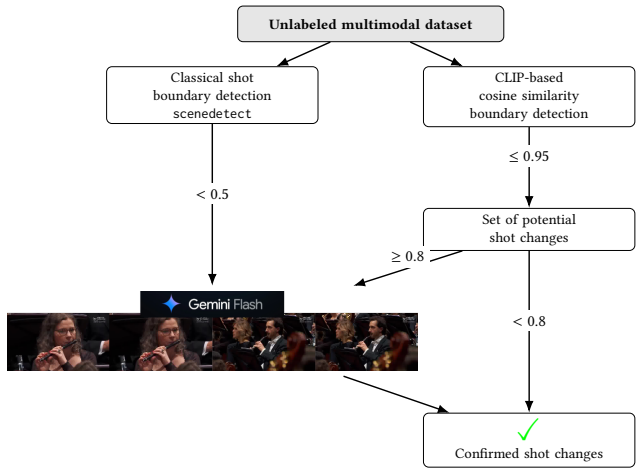
\begin{figure}[!ht]
    \centering
    \resizebox{\columnwidth}{!}{%
        \begin{tikzpicture}[
            node distance=1.2em and 2.5em,
            box/.style={
                rectangle, draw, rounded corners=3pt,
                align=center, font=\small,
                text width=0.22\textwidth,
                minimum height=2.2em
            },
            widebox/.style={
                rectangle, draw, rounded corners=3pt,
                align=center, font=\small,
                text width=0.28\textwidth,
                minimum height=2.2em
            },
            arrow/.style={-Latex, thick}
        ]

            \node[widebox, fill=gray!20] (raw) {\textbf{Unlabeled multimodal dataset}};

            \node[box, below left=1.5em and -5em of raw] (classical) {Classical shot\\ boundary detection\\ \texttt{scenedetect}};
            \node[box, below right=1.5em and -5em of raw] (clip) {CLIP-based\\ cosine similarity \\ boundary detection};


            \node[box, below=5em of clip] (set2) {Set of potential\\ shot changes};
            \node[box, below=10em and -5.5em of set2] (confirmed) {{\Huge\textcolor{green}{\textbf{$\checkmark$}}}\\Confirmed shot changes};

            \node[below=10em of classical] (gemini) {\includegraphics[width=0.45\textwidth]{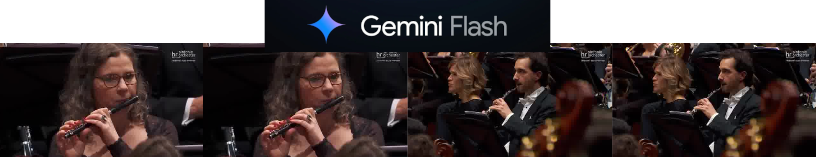}};

            \draw[arrow] (raw) -- (classical);
            \draw[arrow] (raw) -- (clip);
            \draw[arrow] (classical) -- node[midway, above, fill=white]{$<$ 0.5} (gemini);
            \draw[arrow] (clip) -- node[midway, above, fill=white]{$\leq$ 0.95} (set2);
            \draw[arrow] (set2) -- node[midway, above, fill=white]{$<$ 0.8} (confirmed);
            \draw[arrow] (set2) -- node[midway, above, fill=white]{$\geq$ 0.8} (gemini);
            \draw[arrow] (gemini) -- (confirmed);

        \end{tikzpicture}
    }
    \caption{
        Pipeline for shot boundary detection combining classical and semantic approaches.  A raw video is first processed using the classical shot boundary detector \texttt{scenedetect}, which yields a set of potential cut locations. In parallel, CLIP embeddings are computed for each frame and compared using cosine similarity to estimate content changes between frames. The output from both streams is merged to confirm true shot transitions. Finally, the confirmed shots are passed through a Gemini 1.5 Flash-based model for downstream tasks such as captioning.
    }
    \label{fig:cut_detector}
\end{figure}

 On the other hand, for the \emph{temporal segmentation task} (\emph{when to cut}), we did not want to solely depend on the pixel-wise differences, and we wanted to take advantage of transformer-based architectures, as author have shown that they contain useful information for semantic segmentation of videos~\cite{caravaca2025vclipper,lee2024pseudo}. We started by detecting scenes changes, but now using a CLIP-based model plus a cosine similarity thresholding between subsequent frames. We discarded frames with a cosine similarity above 0.95, and automatically accepted those cuts who presented less than 0.8 cosine similarity. For the middle range, we assumed a diffuse threshold and seek for a confirmation of the cut by using a Gemini 1.5 Flash model~\cite{gemini1.5} to verify the presence of a shot change. The Gemini model was prompted with five surrounding frames, asking it to confirm the cut by providing a joined image of the shots. This approach allowed us to leverage the semantic understanding of the Gemini model to validate potential cuts that were not clearly defined by pixel differences alone. Lastly, in order to retrieve more meaningful cuts, we used the \texttt{scenedetect} library to detect cuts in the original high-resolution videos--using its \emph{adaptive detector} with a default threshold of 0.5, which compares the average color between frames of the HSV color space--,and all of these cuts where submitted to the Gemini 1.5 Flash model to verify the presence of a shot change. Details can be seen in~\autoref{fig:cut_detector}.

Finally, the dataset was labelled differently for the two tasks, and subsequently the train-validation-test splits were performed differently, but in similar fashion, over the two pseudo-labeled datasets. In all cases, we ensured that no scene cut from a video in the test set was used in the train or validation sets, to avoid data leakage and ensure that the models were evaluated on unseen data. The details of the pseudo-labeling and splits were as follows:

\textbf{Temporal segmentation task:} For the \emph{when to cut} task, we decided to split the videos in the dataset into segments of 4 seconds, overlapped 2 seconds. For each of these video segments, we observed if a cut was detected early in with the before-mentioned pipeline, and if so, we assigned a label of 1 to the segment, indicating that it contained a cut. If no cut was detected, the segment was assigned a label of 0, indicating that it did not contain a cut. Furthermore, another scalar label was assigned to each of the 4 seconds segments, related with the time elapsed since the last cut. The label was defined as follows:

        \begin{equation}
            l_{seg} = \frac{(\alpha_\text{seg} - \alpha_{\text{shot}}) - {\bar{t}_{\text{scene}}}}{\sigma_{\text{scene}}},
            \label{eq:time_label}
        \end{equation}

    where \(\alpha_\text{seg}\) and \(\alpha_\text{shot}\) are the indexes of the samples where the processed segment starts and the previous change of shot occurs respectively, while \({\bar{t}_{\text{scene}}}\) and \({\sigma_{\text{scene}}}\) are the mean and standard deviation distance between two shot changes (in number of samples) of the whole training set.


\textbf{Spatial selection task:} To segment videos into visually coherent units (e.g., pseudo camera shots), we used the CLIP ViT-B/32 model to extract high-level semantic features from each frame, capturing both low- and high-level visual information~\cite{lin2024videogenic,caravaca2025vclipper,lee2024pseudo}. These features were reduced via PCA while preserving at least 66\% of the variance, and then clustered using K-Means. The number of clusters was automatically selected by maximizing the silhouette score. Only videos with at least 6 distinct clusters were retained for the subsequent analysis, ensuring sufficient visual diversity.

Cluster labels were then used to construct training pairs for a frame transition prediction task. For each frame (the \emph{anchor}), the next frame from the actual video stream served as a \emph{positive} example. To generate hard negatives, we sampled \textbf{9 distractor frames} from the same video that did not belong to the clusters of either the anchor or the positive. This yielded ten pairs per anchor, of the form $(\text{anchor}, \text{candidate}, \text{label})$ with $\text{label} \in \{0, 1\}$.

\subsection{Model architectures}\label{subsec:model-architectures}

\subsubsection{When to Cut? Temporal Segmentation Task}\hfill \\

For this task, we proposed two variants of a shot boundary detection model: a unimodal model based solely on audio and timing information, and a multimodal extension that also incorporated visual data. Both models aimed to predict whether a scene cut occurred within a given four-second audio segment, producing a binary output interpreted as the probability of a transition.

\begin{figure*}[ht!]
    \centering

    \begin{subfigure}[b]{0.20\textwidth}
        \centering
        \caption{}
        \includegraphics[width=\textwidth]{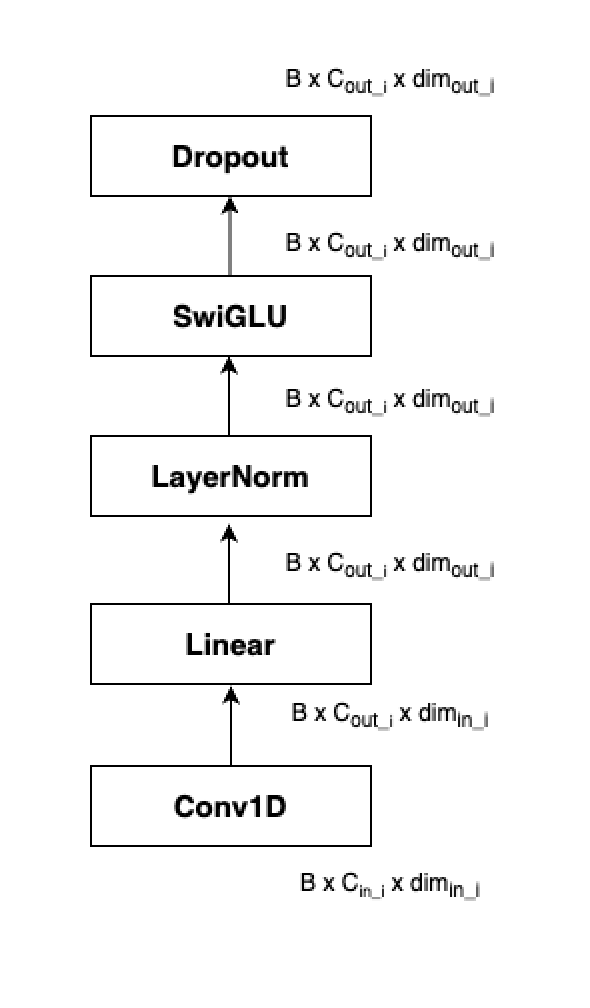}
        \label{fig:cnn_block}
    \end{subfigure}
    \begin{subfigure}[b]{0.65\textwidth}
        \centering
        \caption{}
        \includegraphics[width=\textwidth]{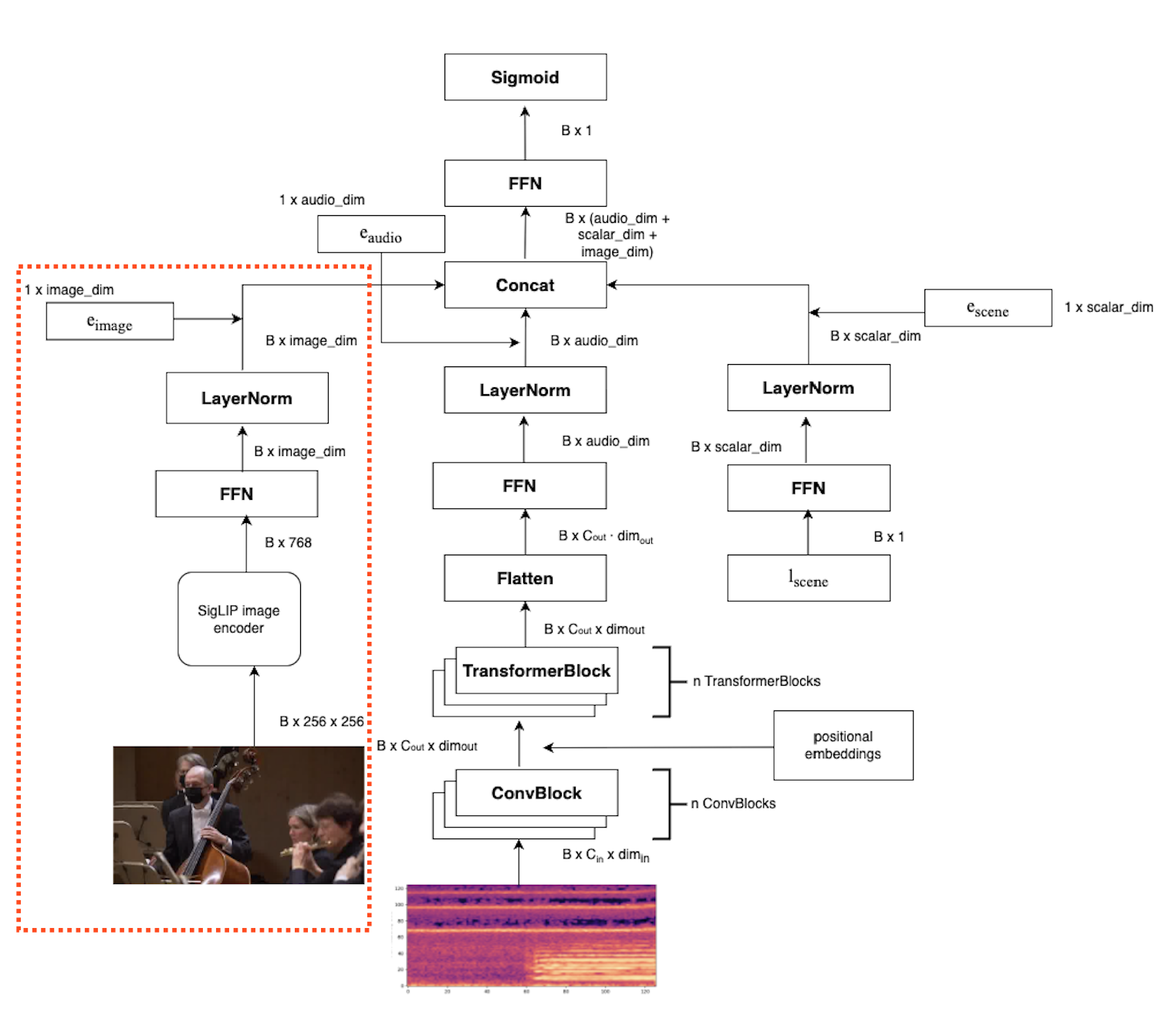}
        \label{fig:multimodal}
    \end{subfigure}
    \caption{\textit{Overview of the temporal segmentation model.} (a) Internal structure of a convolutional block. (b) Architecture of the full multimodal model, which processes audio, video and and time features into a single embedding. The audio input is a log-mel spectrogram, while the time input is a scalar value representing the time elapsed since the last cut. The visual input is an image embedding from a pretrained vision model. The final output is a probability of a scene cut occurring within the given segment. The unimodal model is a simplified version of this architecture, the dashed red rectangle indicates the components that are not present in the unimodal model.}
    \label{fig:multimodal-cnnblock}
\end{figure*}

The \textit{unimodal model} took as input two elements: a log-mel spectrogram of shape $128 \times 400$ pixels, representing the audio content, and the scalar value $l_{\text{seg}}$ encoding the time elapsed since the last cut, as before-mentioned. To process the audio input, we designed a stack of convolutional blocks (\autoref{fig:cnn_block}). Each block includes a one-dimensional convolution along the frequency axis, a linear transformation, a layer normalization, and a SwiGLU activation~\cite{glu_variants}. The convolutional layers extract frame-wise spectral features, while the linear projections aggregate temporal information. To enable sequence modeling, we added fixed, non-trainable sinusoidal positional embeddings~\cite{attention}, and fed the resulting representations into transformer layers. These layers capture longer-term dependencies across the input sequence. The output sequence was flattened and passed through a feed-forward network that projects the result into a fixed-size audio embedding $\mathbf{f}_{\text{audio}}$.

In parallel, the model defines a scalar input $l_{\text{seg}}$ using a small feed-forward network that projects it into a vector space of dimension $\texttt{time\_dim} = \texttt{audio\_dim} / 2$. A learnable embedding vector $\mathbf{e}_{\text{scene}}$ was added to this projection to help the model distinguish between modalities. The resulting vector $\mathbf{f}_{\text{scene}}$ is normalized to prevent scale imbalances. We then defined the model to concatenate the normalized embeddings $\mathbf{f}_{\text{audio}}$ and $\mathbf{f}_{\text{scene}}$ and to pass the result through a final feed-forward network, which concludes with a sigmoid activation. This output represents the model's predicted probability of a scene cut. An overview of the full model architecture can be seen in~\autoref{fig:multimodal}.

For the \textit{multimodal model}, we extended this setup by introducing a third input: the most recent video frame preceding the audio segment, available due to the dataset's 5 FPS sampling rate. This image was encoded using the early layers of a SigLIP2 vision model~\cite{tschannen2025siglip2multilingualvisionlanguage}, including its initial convolutions, embeddings, and attention blocks. A custom linear layer mapps the visual output to the same embedding space as the audio vector. To ensure meaningful semantic representations in the classical music domain --e.g., distinguishing between instruments or performers-- we fine-tuned the selected SigLIP2 layers.


\subsubsection{How to cut? Spatial selection task}\hfill \\

For this task, we proposed a modern approach to Jimenez et al.~\cite{jimenez2022automated} by replacing their ResNet-based backbone with a CLIP-based encoder to improve the alignment between visual features and high-level semantics. This approach consists on using a pretrained vision model to extract visual features from the video frames, and then using these features to select the best shot from a set of candidates, using a matching module to compute relevance scores between the anchor and each candidate. Regarding the pretrained model, we used the CLIP ViT-B/32 model~\cite{radford2021learning}, which embeds each input image into a 512-dimensional latent space.

The matching module was defined as an attention-based mechanism to compute relevance scores between the anchor and each candidate. In these kind of schemes, let $a \in \mathbb{R}^d$ be the embedding of the anchor frame and $C \in \mathbb{R}^{N \times d}$ the embeddings of $N=10$ candidates. Both the anchor and candidates are first projected into a shared latent space via independent linear layers followed by a $\tanh$ activation. Then, the similarity score between the anchor and each candidate is computed using scaled dot-product attention:

\begin{equation}
    s_i = \frac{(W_k c_i)^\top (W_q a)}{\sqrt{d}},
    \label{eq:attention_score}
\end{equation}

where $W_k$ and $W_q$ are learnable projection matrices, and $d$ is the dimension of the projected space. This results in a score vector $s \in \mathbb{R}^{10}$, which is interpreted as unnormalized logits over the candidates. Lastly, the logits are passed through a softmax activation to obtain a probability distribution over the candidates, which is then used to compute the loss. This architecture provides a balance between semantic understanding and computational efficiency. The use of CLIP embeddings enhances scene understanding, while the lightweight attention-based scorer allows for fast inference.

\section{Results}\label{sec:results}

\subsection{When to cut?}\label{subsec:when-to-cut?}

Since the problem we addressed has not been previously studied in the literature in this form, there are no existing state-of-the-art methods or public benchmarks to serve as a basis for comparison. Consequently, we defined two statistical baselines inspired by prior work on automated video editing with data-driven strategies~\cite{jimenez2022automated}, but not exactly the same. Instead of sampling from the actual distribution from the dataset, we fitted two statistical baseline models to the data, that were the Poisson and the exponential distributions. For the \emph{Poisson-based method}, the expected number of scene changes in a fixed time interval, as the distribution is defined as:

\begin{equation}
    P_{\text{Poi}}(x=k) = \frac{\lambda^k e^{-\lambda}}{k!},
\end{equation}

where \( \lambda \in \mathbb{R}^+ \) represents the expected number of shot changes per time window. We estimated \( \lambda \) as the ratio between the input window duration and the average scene duration in the training set:

\begin{equation}
    \lambda = \frac{t_{\text{window}}}{\bar{t}_{\text{scene}}}
\end{equation}

where, \( t_{\text{window}} \) is the length of the input video segment and \( \bar{t}_{\text{scene}} \) is the average scene duration computed from the training set, same as~\autoref{eq:time_label}. This approach assumes that the number of scene changes follows a Poisson distribution, which is suitable for modeling discrete events occurring independently over time. From this distribution, we sampled a value \( x \sim P_\text{Poi}(\lambda) \). If \( x > 0 \), the sample was labeled as positive, indicating the presence of a scene change in that interval.

The \emph{Exponential-based method}, in contrast, models the distribution of time intervals between two consecutive cut events. The exponential distribution is defined by:
\begin{equation}
    P_{\text{Exp}}(x=k) = \alpha \cdot e^{-\alpha k},
\end{equation}

where \( \alpha^{-1} = \bar{t}_{\text{scene}} \). We sampled a scene duration \( x \sim P_\text{Exp}(\alpha) \) and compared it to the actual length \( t_{\text{scene}} \) of the evaluated input. If \( t_{\text{scene}} < x \), the sample was classified as containing a cut.

These baselines, while not learning-based, provide meaningful benchmarks rooted in the statistical regularities observed in professionally edited material. Additionally, we derived a probabilistic output  from both models by computing the ROC-AUC curve to be compared with our models. For the Poisson model, we estimated the score as \( p = 1 - \text{CDF}(0) \), where \(\text{CDF}(x)\) denotes the cumulative distribution function. For the Exponential model, we used \( p = \text{CDF}(t_{\text{scene}}) \). These scores enabled a fair comparison with the neural models, whose outputs are naturally probabilistic.

Regarding our deep-learning models, we trained both the unimodal and multimodal architectures using the AdamW optimizer with a learning rate of \(5 \times 10^{-4}\) and a weight decay of 0.01. The training process involved minimizing the binary cross-entropy loss function, which is suitable for binary classification tasks. We employed a linear learning rate scheduler with a warm-up phase of 5 epochs to stabilize training in the initial stages. As this task presented a significant class imbalance, as more positive samples were present than negative ones, we implemented a augmentation scheme using SpecAugment~\cite{park2020specaugment} to even the number of positive and negative samples. This augmentation involved applying time and frequency masking to the log-mel spectrograms, as well as adding noise sampled from a uniform distribution. The splits for this task were performed as follows: 15 videos where reserved solely for testing, while the data from the remaining 85 videos was split into training and validation sets, where from the 100\% of the training-validation data, around 80\% was used for training and 20\% for validation. The training-validation split was performed in a stratified manner, ensuring that the distribution of positive and negative samples was preserved in both sets.

The results of the classification task were evaluated in terms of precision, recall, F1-score, accuracy, and ROC-AUC score for both models and baselines. Table~\ref{tab:classification_results} summarizes the performance of the baseline statistical methods and the proposed unimodal and multimodal models on both the validation and test sets. In the validation set, the Exponential-based method outperformed the Poisson-based method in terms of F1-score. This result aligned with expectations, as the exponential distribution is better suited for modeling the time intervals between events, making it more appropriate for this task. However, both statistical baselines were clearly outperformed by the proposed learning-based models. The unimodal and multimodal models achieved substantial improvements across all metrics, with the unimodal model reaching the highest recall and F1-score, while the multimodal model attained the best precision.

\begin{table}[ht]
    \centering
    \caption{Comparison of the baseline statistical methods and the proposed unimodal and multimodal models on the validation and test sets. The results are shown in terms of precision, recall, F1-score, and accuracy. The best results for each metric are highlighted in bold.}
    \label{tab:classification_results}
    \resizebox{\columnwidth}{!}{%
        \begin{tabular}{l|c|c|c|c}
            \toprule
            {\small \textbf{Model}} &
                {\scriptsize \textbf{Precision (\%)}} &
                {\scriptsize \textbf{Recall (\%)}} &
                {\scriptsize \textbf{F1-score (\%)}} &
                {\scriptsize \textbf{Accuracy (\%)}} \\
            \midrule
            \multicolumn{5}{l}{\textit{Validation set}} \\
            \midrule
            Baseline (Exp)   & 47.66 & 68.34 & 56.16 & 48.25 \\
            Baseline (Pois)  & 48.25 & 28.31 & 35.69 & 50.55 \\
            Unimodal model   & 61.39 & \textbf{71.42} & \textbf{66.03} & \textbf{64.38} \\
            Multimodal model & \textbf{61.52} & 68.45 & 64.80 & 63.96 \\
            \midrule
            \multicolumn{5}{l}{\textit{Test set}} \\
            \midrule
            Baseline (Exp)   & 50.22 & \textbf{70.14} & 58.53 & 49.63 \\
            Baseline (Pois)  & 50.18 & 28.31 & 36.20 & 49.42 \\
            Unimodal model   & \textbf{61.30} & 67.89 & \textbf{64.43} & \textbf{62.01} \\
            Multimodal model & 61.14 & 63.57 & 62.33 & 61.06 \\
            \bottomrule
        \end{tabular}
    }
\end{table}

\begin{figure}[htbp]
    \centering
    \begin{subfigure}[b]{0.85\linewidth}
        \centering
        \caption{}
        \includegraphics[width=\linewidth,trim={0 0 0 14mm},clip]{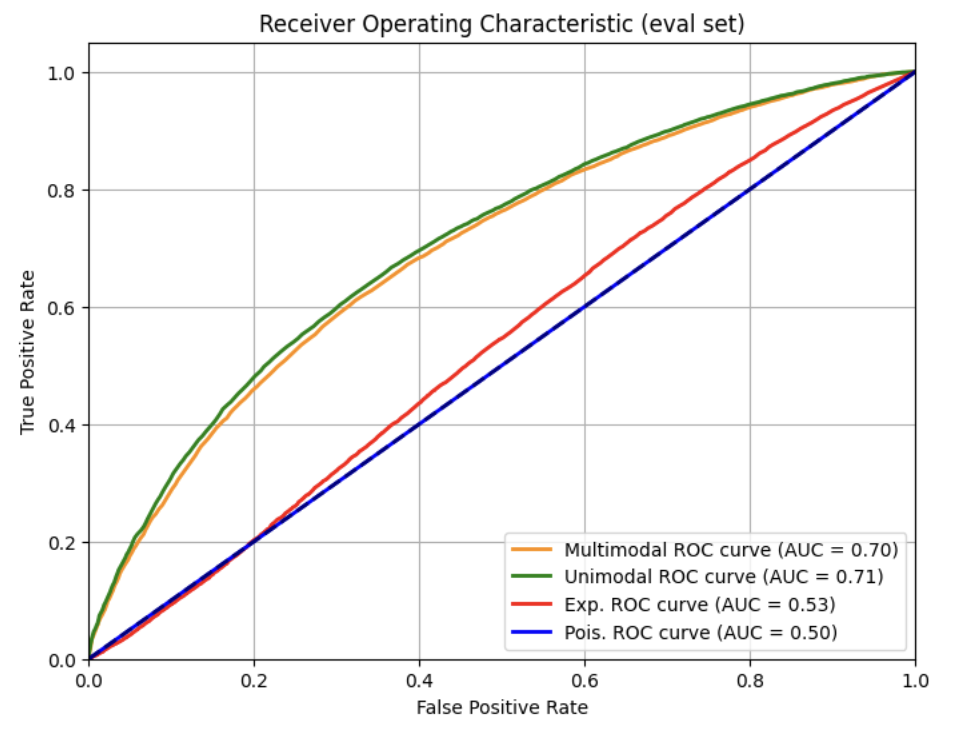}
    \end{subfigure}


    \begin{subfigure}[b]{0.85\linewidth}
        \centering
        \caption{}
        \includegraphics[width=\linewidth,trim={0 0 0 15mm},clip]{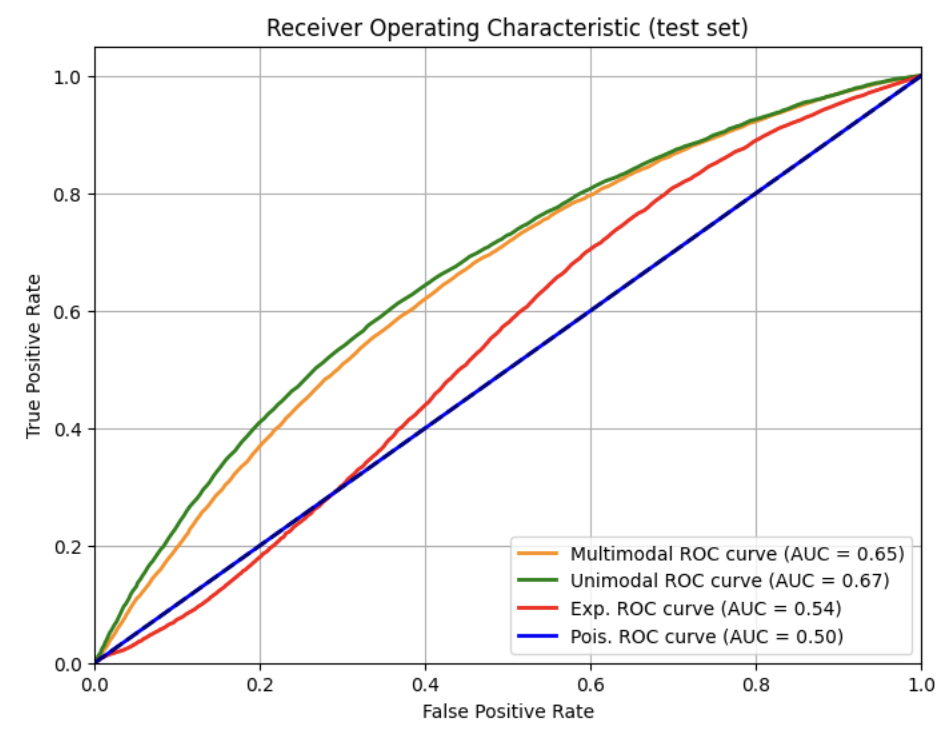}
    \end{subfigure}

    \caption{ROC curves for the classification task. (a) Validation set: the unimodal and multimodal models showed clearly superior performance over the statistical baselines across all thresholds. (b) Test set: although performance slightly decreased, the models retained good ranking capabilities.}
    \label{fig:roc_auc_subfigs}
\end{figure}

Figure~\ref{fig:roc_auc_subfigs} illustrates the ROC curves for both the validation and test sets. In the validation set (Figure~\ref{fig:roc_auc_subfigs}a), the curves of the learning-based models consistently outperformed those of the statistical baselines across all thresholds, with a substantial margin between the curves. On the test set (Figure~\ref{fig:roc_auc_subfigs}b), performance slightly declined, especially in the recall and F1-score values, but the ROC curves maintained their relative shapes and order. This result suggested that the models preserved their ranking capabilities and generalization potential when applied to unseen videos, which may feature different musical styles and structures than those present during training.

\subsection{How to cut?}\label{subsec:how-to-cut?}

Regarding the \emph{how to cut} task, we defined several baselines to compare against our proposed model. Basically, we introduced similar baselines as related works, first a random baseline was considered, e.g. selecting a random candidate from the ten available options, this is 1/10 chance of selecting the correct candidate. Additionally, the same model was considered with older backbones, such as ResNet-50 and Xception~\cite{jimenez2022automated}, which were trained on the same dataset. These models were used to extract visual features from the video frames, which were then used to select the best shot from a set of candidates.

All models where trained using the binary cross-entropy loss applied to the one-hot target vector, where the target vector is a one-hot encoding of the correct candidate. The optimization was performed using the Adam optimizer with a learning rate of \(1 \times 10^{-5}\). The training-validation-test splits were performed in a similar manner as for the temporal segmentation task, but selecting different splits, as the sub-datasets were labelled differently. By construction, and on purpose, this sub-dataset was unbalanced, in the sense that each triplet of data consisted of one anchor frame, one positive candidate (the correct shot) and nine negative candidates (distractors), but each triplet is similar to the others, so the dataset as a whole is balanced, as the real-life problem includes deciding where to cut from 1 to N frame, as per modeled here 10 of them. In this case, 15\% of the videos were reserved for testing, while the remaining 85\% video cuts were split into training and validation sets, 15\% of those videos were used for validation and the rest for training (70\% of the total dataset).

For the evaluation metric we used the top-1 accuracy (Recall@1) and top-3 accuracy (Recall@3) over a 10-way candidate selection task, where the model must correctly identify the next camera shot among ten possible options.~\autoref{tab:shot_prediction_performance} summarizes the results of the shot prediction task using different visual feature extractors. The table shows the Recall@1 and Recall@3 scores for each model, indicating the percentage of times the correct shot was ranked first or within the top three candidates, respectively.

\begin{table}[h]
\centering
\caption{Shot prediction performance using different visual feature extractors. The best results for each metric are highlighted in bold.}
\label{tab:shot_prediction_performance}
\begin{tabular}{lcc}
\toprule
\textbf{Model} & \textbf{Recall@1 (\%)} & \textbf{Recall@3 (\%)} \\
\midrule
Random & 10.00 & 30.00 \\
ResNet-50 & 11.18 & 30.08 \\
Xception & 25.08 & 48.89 \\
CLIP ViT-B/32 & \textbf{28.49} & \textbf{51.97} \\
\bottomrule
\end{tabular}
\end{table}

While all models perform above random chance (10\% Recall@1 and 30\% Recall@3), the results indicate that semantically rich feature extractors like CLIP and Xxception yield significantly better performance on both top-1 and top-3 accuracy metrics, as ResNet-50, which is a really old model, achieved only 11.18\% Recall@1 and 30.08\% Recall@3, which the same results as random selection. Xception, a more recent architecture, improves the performance to 25.08\% Recall@1 and 48.89\% Recall@3, but still falls short of the CLIP ViT-B/32 model, which achieves 28.49\% Recall@1 and 51.97\% Recall@3. This demonstrates the effectiveness of using CLIP embeddings for capturing high-level visual semantics relevant to shot transitions in videos.

\begin{figure*}[t]
    \centering
    \begin{subfigure}[b]{\textwidth}
        \centering
        \caption{}
        \includegraphics[width=\textwidth]{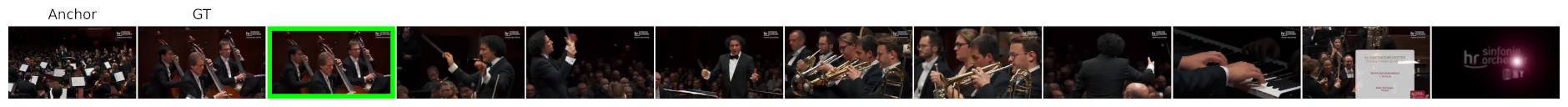}
        \label{fig:Top1}
    \end{subfigure}
    \vspace{0.5em}

    \begin{subfigure}[b]{\textwidth}
        \centering
        \caption{}
        \includegraphics[width=\textwidth]{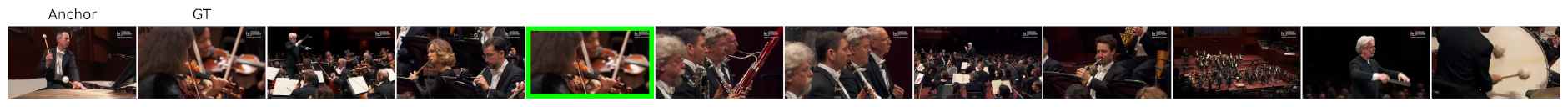}
        \label{fig:Top3}
    \end{subfigure}
    \vspace{0.5em}

    \begin{subfigure}[b]{\textwidth}
        \centering
        \caption{}
        \includegraphics[width=\textwidth]{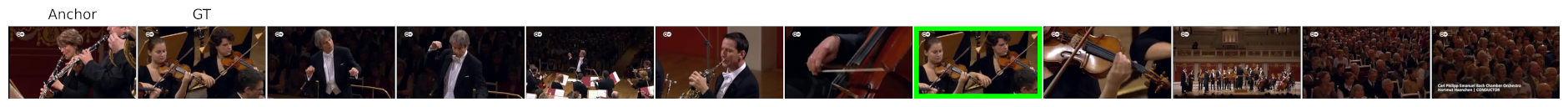}
        \label{fig:Top6}
    \end{subfigure}

    \caption{Prediction examples ranked by confidence: (a) the model’s top-1 predicted candidate matches the ground truth (GT); (b) the GT appears within the top 3 predictions but not first; (c) the GT is within the top 6 predictions but outside the top 3, indicating lower model confidence.}
    \label{fig:TopN}
\end{figure*}

On a more qualitative note, we present three examples of model predictions in \autoref{fig:TopN}. Each example consists of an anchor frame (the actual shot frame being queried), the ground truth (GT) shot, and the top 10 candidate predictions ranked by the model's predicted probabilities. The prediction framed in green corresponds to the GT shot. The examples illustrate different scenarios of model performance:

\begin{itemize}
    \item \textbf{Top1 (\autoref{fig:Top1}):} The model correctly identifies the ground truth frame as the top-1 prediction, demonstrating an ideal case where the highest-ranked candidate matches the actual shot transition.
    \item \textbf{Top3 (\autoref{fig:Top3}):} The ground truth frame appears within the top 3 predicted candidates, indicating that while the model's highest confidence prediction was incorrect, it still ranks the correct shot transition near the top.
    \item \textbf{Top6 (\autoref{fig:Top6}):} The ground truth frame is not within the top 3 but appears within the top 6 candidates, reflecting cases where the model's performance is weaker and the correct shot transition is less confidently predicted.
\end{itemize}

\section{Conclusions and Discussion}\label{sec:conclusions-and-discussion}


In this work, we proposed a multimodal approach for the automated editing of classical music concert videos, addressing both the when to cut and how to cut sub-tasks through modern deep learning strategies. Our method integrates log-mel spectrograms, scalar temporal features, and optionally visual embeddings extracted using CLIP, all processed through a lightweight convolutional-transformer architecture. To support this architecture, we constructed a pseudo-labeled dataset of 100 classical concerts, using a hybrid shot detection pipeline that combines classical heuristics, CLIP-based semantic similarity, and LLM-based confirmation via Gemini. This allowed us to label transitions with high reliability while preserving the semantic richness of the original content.
Experimental results showed that our unimodal model achieved an accuracy of 62.01

Looking forward, several promising research directions emerge from this work. First, the temporal segmentation component, currently formulated as a binary classification task, could be naturally extended into a continuous regression framework that predicts the precise timestamp of the next optimal cut. Such a formulation would allow finer-grained modeling of transitions and improve temporal coherence. Second, the audio modality, currently based on low-level time-frequency representations, could benefit from higher-level semantic or affective cues. Inspired by recent work on musical emotion and aesthetic perception~\cite{xu2025aesthetic}, future models could learn to synchronize cuts with emotionally salient moments in the audio stream. Similarly, on the visual side, emotion recognition could enhance the spatial selection process by prioritizing shots that highlight expressive content—such as facial expressions, instrumental gestures, or audience reactions—further aligning the editing decisions with the emotional arc of the performance.

Despite the consistent improvements over strong baselines, the absolute performance values, 62.01\% accuracy for the when to cut model and 28.49\% Recall@1 for the how to cut model highlight limitations in robustness for production-grade deployment. One key contributing factor lies in the evaluation strategy itself: in the how to cut task, the model is penalized for selecting a candidate view that, while different from the ground truth, presents highly similar semantic content (e.g., adjacent camera angles of the same performer). These visually or contextually equivalent frames often share similar feature maps or embedding representations, especially when extracted via CLIP. Thus, choosing Camera 2 instead of the annotated Camera 1, despite showing nearly identical visual information, is currently counted as an error, which underrepresents the practical adequacy of the model's decision.

To address this, future work will explore similarity-aware evaluation protocols. Specifically, we plan to introduce a soft matching scheme that assigns partial credit to predictions based on their semantic proximity to the ground truth, possibly through embedding-space distance metrics. We also envision incorporating a dynamic camera-weighting mechanism, where each candidate shot is assigned a learned "view redundancy" score that modulates its impact on the final loss and evaluation, encouraging diversity when beneficial, and penalizing only truly divergent transitions.

In terms of dataset generalizability, the current study focused solely on classical music performances in order to maintain high consistency across visual style, camera framing, and audio structure. This homogeneity allowed us to reduce confounding variables and isolate core editing behaviors. However, this domain-specific focus limits the applicability of the current models to more dynamic or heterogeneous settings. In future work, we plan to extend our approach to domains where automated editing is highly demanded—such as sports broadcasts, stage plays, or public speaking—by curating new datasets and leveraging existing public \cite{rao2022temporalcontextualtransformermulticamera} resources designed for video editing tasks. This will not only help benchmark our models across a wider range of use cases but also uncover domain-specific adaptations needed for broader applicability.

Moreover, we foresee the possibility of introducing user-in-the-loop or prompt-based editing schemes, where high-level instructions guide the system toward particular visual styles or musical interpretations. Finally, while our current formulation focuses on local decisions, future approaches may incorporate long-term temporal modeling or memory mechanisms, enabling globally coherent editing strategies that unfold across entire movements or full-length performances.

Altogether, this work contributes to bridging the gap between low-level vision tasks and high-level editorial decisions in the context of multicamera concert editing. By leveraging multimodal signals and modern deep learning architectures, we lay the foundation for more expressive, adaptive, and semantically grounded automated editing systems.

\newpage
\begin{acks}
  Authors acknowledge the funding support from grant  \textbf{PID2022-138721NB-I00}, funded by the MCIN (Spain). I. Benito-Altamirano acknowledges the support from grant \textbf{2025-DI-00035} funded by the Industrial Doctorates Plan from AGAUR (Catalonia, Spain). 
\end{acks}

\bibliographystyle{ACM-Reference-Format}
\balance
\bibliography{main}


\begin{thebibliography}{23}


\ifx \showCODEN    \undefined \def \showCODEN     #1{\unskip}     \fi
\ifx \showISBNx    \undefined \def \showISBNx     #1{\unskip}     \fi
\ifx \showISBNxiii \undefined \def \showISBNxiii  #1{\unskip}     \fi
\ifx \showISSN     \undefined \def \showISSN      #1{\unskip}     \fi
\ifx \showLCCN     \undefined \def \showLCCN      #1{\unskip}     \fi
\ifx \shownote     \undefined \def \shownote      #1{#1}          \fi
\ifx \showarticletitle \undefined \def \showarticletitle #1{#1}   \fi
\ifx \showURL      \undefined \def \showURL       {\relax}        \fi
\providecommand\bibfield[2]{#2}
\providecommand\bibinfo[2]{#2}
\providecommand\natexlab[1]{#1}
\providecommand\showeprint[2][]{arXiv:#2}

\bibitem[Amiriparian et~al\mbox{.}(2017)]%
        {amiriparian2017snore}
\bibfield{author}{\bibinfo{person}{Shahin Amiriparian},
  \bibinfo{person}{Maurice Gerczuk}, \bibinfo{person}{Sandra Ottl},
  \bibinfo{person}{Nicholas Cummins}, \bibinfo{person}{Michael Freitag},
  \bibinfo{person}{Sergey Pugachevskiy}, \bibinfo{person}{Alice Baird}, {and}
  \bibinfo{person}{Bj{\"o}rn Schuller}.} \bibinfo{year}{2017}\natexlab{}.
\newblock \showarticletitle{Snore sound classification using image-based deep
  spectrum features}.
\newblock  (\bibinfo{year}{2017}).
\newblock


\bibitem[Caravaca-M{\"u}ller et~al\mbox{.}(2025)]%
        {caravaca2025vclipper}
\bibfield{author}{\bibinfo{person}{Oriol Caravaca-M{\"u}ller},
  \bibinfo{person}{Joan Llobera}, \bibinfo{person}{Carles Ventura}, {and}
  \bibinfo{person}{Ismael Benito-Altamirano}.} \bibinfo{year}{2025}\natexlab{}.
\newblock \showarticletitle{VClipper: Moment Retrieval in Video Streams Using
  Zero-Shot and Context-Aware Foundational Models}. In
  \bibinfo{booktitle}{\emph{International Conference on Advanced Information
  Networking and Applications}}. Springer, \bibinfo{pages}{411--422}.
\newblock


\bibitem[Castellano)(2025)]%
        {scenedetect}
\bibfield{author}{\bibinfo{person}{{Breakthrough}~(Brandon Castellano)}.}
  \bibinfo{year}{2025}\natexlab{}.
\newblock \bibinfo{title}{PySceneDetect: Open-source scene‑cut/shot boundary
  detection tool}.
\newblock \bibinfo{howpublished}{Version 0.6.6,
  \url{https://github.com/Breakthrough/PySceneDetect}}.
\newblock
\newblock
\shownote{Python API and CLI for detecting scene changes using Content,
  Adaptive, Histogram, Hash, and Threshold detectors}.


\bibitem[Del~Fabro and B{\"o}sz{\"o}rmenyi(2013)]%
        {del2013state}
\bibfield{author}{\bibinfo{person}{Manfred Del~Fabro} {and}
  \bibinfo{person}{Laszlo B{\"o}sz{\"o}rmenyi}.}
  \bibinfo{year}{2013}\natexlab{}.
\newblock \showarticletitle{State-of-the-art and future challenges in video
  scene detection: a survey}.
\newblock \bibinfo{journal}{\emph{Multimedia systems}}  \bibinfo{volume}{19}
  (\bibinfo{year}{2013}), \bibinfo{pages}{427--454}.
\newblock


\bibitem[{Gemini Team, Google DeepMind}(2024)]%
        {gemini1.5}
\bibfield{author}{\bibinfo{person}{{Gemini Team, Google DeepMind}}.}
  \bibinfo{year}{2024}\natexlab{}.
\newblock \bibinfo{title}{Gemini 1.5: Unlocking multimodal understanding across
  millions of tokens of context}.
\newblock \bibinfo{howpublished}{arXiv preprint arXiv:2403.05530}.
\newblock
\urldef\tempurl%
\url{https://arxiv.org/abs/2403.05530}
\showURL{%
\tempurl}
\newblock
\shownote{Includes Gemini 1.5 Flash, a distilled variant optimized for
  high-efficiency multimodal inference}.


\bibitem[Henschel et~al\mbox{.}(2025)]%
        {henschel2025streamingt2v}
\bibfield{author}{\bibinfo{person}{Roberto Henschel}, \bibinfo{person}{Levon
  Khachatryan}, \bibinfo{person}{Hayk Poghosyan}, \bibinfo{person}{Daniil
  Hayrapetyan}, \bibinfo{person}{Vahram Tadevosyan}, \bibinfo{person}{Zhangyang
  Wang}, \bibinfo{person}{Shant Navasardyan}, {and} \bibinfo{person}{Humphrey
  Shi}.} \bibinfo{year}{2025}\natexlab{}.
\newblock \showarticletitle{Streamingt2v: Consistent, dynamic, and extendable
  long video generation from text}. In \bibinfo{booktitle}{\emph{Proceedings of
  the Computer Vision and Pattern Recognition Conference}}.
  \bibinfo{pages}{2568--2577}.
\newblock


\bibitem[Jim{\'e}nez et~al\mbox{.}(2022)]%
        {jimenez2022automated}
\bibfield{author}{\bibinfo{person}{Albert Jim{\'e}nez},
  \bibinfo{person}{Llu{\'\i}s G{\'o}mez}, {and} \bibinfo{person}{Joan
  Llobera}.} \bibinfo{year}{2022}\natexlab{}.
\newblock \showarticletitle{Automated Video Edition for Synchronized Mobile
  Recordings of Concerts.}. In \bibinfo{booktitle}{\emph{VISIGRAPP (4:
  VISAPP)}}. \bibinfo{pages}{941--948}.
\newblock


\bibitem[Lee et~al\mbox{.}(2024)]%
        {lee2024pseudo}
\bibfield{author}{\bibinfo{person}{Kuan-Ying Lee}, \bibinfo{person}{Qian Zhou},
  {and} \bibinfo{person}{Klara Nahrstedt}.} \bibinfo{year}{2024}\natexlab{}.
\newblock \showarticletitle{Pseudo Dataset Generation for Out-of-domain
  Multi-Camera View Recommendation}. In \bibinfo{booktitle}{\emph{2024 IEEE
  International Conference on Visual Communications and Image Processing
  (VCIP)}}. IEEE, \bibinfo{pages}{1--5}.
\newblock


\bibitem[Lin et~al\mbox{.}(2024)]%
        {lin2024videogenic}
\bibfield{author}{\bibinfo{person}{David Chuan-En Lin}, \bibinfo{person}{Fabian
  Caba~Heilbron}, \bibinfo{person}{Joon-Young Lee}, \bibinfo{person}{Oliver
  Wang}, {and} \bibinfo{person}{Nikolas Martelaro}.}
  \bibinfo{year}{2024}\natexlab{}.
\newblock \showarticletitle{Videogenic: Identifying Highlight Moments in Videos
  with Professional Photographs as a Prior}. In
  \bibinfo{booktitle}{\emph{Proceedings of the 16th Conference on Creativity \&
  Cognition}}. \bibinfo{pages}{328--346}.
\newblock


\bibitem[Mehta et~al\mbox{.}(2021)]%
        {mehta2021music}
\bibfield{author}{\bibinfo{person}{Jash Mehta}, \bibinfo{person}{Deep Gandhi},
  \bibinfo{person}{Govind Thakur}, {and} \bibinfo{person}{Pratik Kanani}.}
  \bibinfo{year}{2021}\natexlab{}.
\newblock \showarticletitle{Music genre classification using transfer learning
  on log-based mel spectrogram}. In \bibinfo{booktitle}{\emph{2021 5th
  International Conference on Computing Methodologies and Communication
  (ICCMC)}}. IEEE, \bibinfo{pages}{1101--1107}.
\newblock


\bibitem[Mensah et~al\mbox{.}(2025)]%
        {mensah2025deep}
\bibfield{author}{\bibinfo{person}{Samuel~Yaw Mensah}, \bibinfo{person}{Tao
  Zhang}, \bibinfo{person}{Nahid~AI Mahmud}, {and} \bibinfo{person}{Yanzhang
  Geng}.} \bibinfo{year}{2025}\natexlab{}.
\newblock \showarticletitle{Deep Learning-Based Speech Enhancement for Robust
  Sound Classification in Security Systems}.
\newblock \bibinfo{journal}{\emph{Electronics}} \bibinfo{volume}{14},
  \bibinfo{number}{13} (\bibinfo{year}{2025}), \bibinfo{pages}{2643}.
\newblock


\bibitem[open-source~project contributors(2021)]%
        {yt-dlp}
\bibfield{author}{\bibinfo{person}{{yt-dlp} open-source~project contributors}.}
  \bibinfo{year}{2021}\natexlab{}.
\newblock \bibinfo{title}{{yt-dlp}: A feature-rich command-line audio/video
  downloader}.
\newblock \bibinfo{howpublished}{\url{https://github.com/yt-dlp/yt-dlp}}.
\newblock
\newblock
\shownote{Accessed: 2025-07-10}.


\bibitem[Ortega-Beltr{\'a}n et~al\mbox{.}(2024)]%
        {ortega2024better}
\bibfield{author}{\bibinfo{person}{Elena Ortega-Beltr{\'a}n},
  \bibinfo{person}{Josep Cabacas-Maso}, \bibinfo{person}{Ismael
  Benito-Altamirano}, {and} \bibinfo{person}{Carles Ventura}.}
  \bibinfo{year}{2024}\natexlab{}.
\newblock \showarticletitle{Better Spanish Emotion Recognition In-the-wild:
  Bringing Attention to Deep Spectrum Voice Analysis}. In
  \bibinfo{booktitle}{\emph{European Conference on Computer Vision}}. Springer,
  \bibinfo{pages}{335--348}.
\newblock


\bibitem[Park et~al\mbox{.}(2020)]%
        {park2020specaugment}
\bibfield{author}{\bibinfo{person}{Daniel~S Park}, \bibinfo{person}{Yu Zhang},
  \bibinfo{person}{Chung-Cheng Chiu}, \bibinfo{person}{Youzheng Chen},
  \bibinfo{person}{Bo Li}, \bibinfo{person}{William Chan},
  \bibinfo{person}{Quoc~V Le}, {and} \bibinfo{person}{Yonghui Wu}.}
  \bibinfo{year}{2020}\natexlab{}.
\newblock \showarticletitle{Specaugment on large scale datasets}. In
  \bibinfo{booktitle}{\emph{ICASSP 2020-2020 IEEE International Conference on
  Acoustics, Speech and Signal Processing (ICASSP)}}. IEEE,
  \bibinfo{pages}{6879--6883}.
\newblock


\bibitem[Quan et~al\mbox{.}(2024)]%
        {quan2024deep}
\bibfield{author}{\bibinfo{person}{Weize Quan}, \bibinfo{person}{Jiaxi Chen},
  \bibinfo{person}{Yanli Liu}, \bibinfo{person}{Dong-Ming Yan}, {and}
  \bibinfo{person}{Peter Wonka}.} \bibinfo{year}{2024}\natexlab{}.
\newblock \showarticletitle{Deep learning-based image and video inpainting: A
  survey}.
\newblock \bibinfo{journal}{\emph{International Journal of Computer Vision}}
  \bibinfo{volume}{132}, \bibinfo{number}{7} (\bibinfo{year}{2024}),
  \bibinfo{pages}{2367--2400}.
\newblock


\bibitem[Radford et~al\mbox{.}(2021)]%
        {radford2021learning}
\bibfield{author}{\bibinfo{person}{Alec Radford}, \bibinfo{person}{Jong~Wook
  Kim}, \bibinfo{person}{Chris Hallacy}, \bibinfo{person}{Aditya Ramesh},
  \bibinfo{person}{Gabriel Goh}, \bibinfo{person}{Sandhini Agarwal},
  \bibinfo{person}{Girish Sastry}, \bibinfo{person}{Amanda Askell},
  \bibinfo{person}{Pamela Mishkin}, \bibinfo{person}{Jack Clark},
  {et~al\mbox{.}}} \bibinfo{year}{2021}\natexlab{}.
\newblock \showarticletitle{Learning transferable visual models from natural
  language supervision}. In \bibinfo{booktitle}{\emph{International conference
  on machine learning}}. PmLR, \bibinfo{pages}{8748--8763}.
\newblock


\bibitem[Rao et~al\mbox{.}(2022)]%
        {rao2022temporalcontextualtransformermulticamera}
\bibfield{author}{\bibinfo{person}{Anyi Rao}, \bibinfo{person}{Xuekun Jiang},
  \bibinfo{person}{Sichen Wang}, \bibinfo{person}{Yuwei Guo},
  \bibinfo{person}{Zihao Liu}, \bibinfo{person}{Bo Dai}, \bibinfo{person}{Long
  Pang}, \bibinfo{person}{Xiaoyu Wu}, \bibinfo{person}{Dahua Lin}, {and}
  \bibinfo{person}{Libiao Jin}.} \bibinfo{year}{2022}\natexlab{}.
\newblock \bibinfo{title}{Temporal and Contextual Transformer for Multi-Camera
  Editing of TV Shows}.
\newblock
\showeprint[arxiv]{2210.08737}~[cs.CV]
\urldef\tempurl%
\url{https://arxiv.org/abs/2210.08737}
\showURL{%
\tempurl}


\bibitem[Shazeer(2020)]%
        {glu_variants}
\bibfield{author}{\bibinfo{person}{Noam Shazeer}.}
  \bibinfo{year}{2020}\natexlab{}.
\newblock \showarticletitle{{GLU} Variants Improve Transformer}.
\newblock \bibinfo{journal}{\emph{CoRR}}  \bibinfo{volume}{abs/2002.05202}
  (\bibinfo{year}{2020}).
\newblock
\showeprint[arXiv]{2002.05202}
\urldef\tempurl%
\url{https://arxiv.org/abs/2002.05202}
\showURL{%
\tempurl}


\bibitem[Tschannen et~al\mbox{.}(2025)]%
        {tschannen2025siglip2multilingualvisionlanguage}
\bibfield{author}{\bibinfo{person}{Michael Tschannen}, \bibinfo{person}{Alexey
  Gritsenko}, \bibinfo{person}{Xiao Wang}, \bibinfo{person}{Muhammad~Ferjad
  Naeem}, \bibinfo{person}{Ibrahim Alabdulmohsin}, \bibinfo{person}{Nikhil
  Parthasarathy}, \bibinfo{person}{Talfan Evans}, \bibinfo{person}{Lucas
  Beyer}, \bibinfo{person}{Ye Xia}, \bibinfo{person}{Basil Mustafa},
  \bibinfo{person}{Olivier Hénaff}, \bibinfo{person}{Jeremiah Harmsen},
  \bibinfo{person}{Andreas Steiner}, {and} \bibinfo{person}{Xiaohua Zhai}.}
  \bibinfo{year}{2025}\natexlab{}.
\newblock \bibinfo{title}{SigLIP 2: Multilingual Vision-Language Encoders with
  Improved Semantic Understanding, Localization, and Dense Features}.
\newblock
\showeprint[arxiv]{2502.14786}~[cs.CV]
\urldef\tempurl%
\url{https://arxiv.org/abs/2502.14786}
\showURL{%
\tempurl}


\bibitem[Vaswani et~al\mbox{.}(2017)]%
        {attention}
\bibfield{author}{\bibinfo{person}{Ashish Vaswani}, \bibinfo{person}{Noam
  Shazeer}, \bibinfo{person}{Niki Parmar}, \bibinfo{person}{Jakob Uszkoreit},
  \bibinfo{person}{Llion Jones}, \bibinfo{person}{Aidan~N. Gomez},
  \bibinfo{person}{Lukasz Kaiser}, {and} \bibinfo{person}{Illia Polosukhin}.}
  \bibinfo{year}{2017}\natexlab{}.
\newblock \showarticletitle{Attention Is All You Need}.
\newblock \bibinfo{journal}{\emph{CoRR}}  \bibinfo{volume}{abs/1706.03762}
  (\bibinfo{year}{2017}).
\newblock
\urldef\tempurl%
\url{http://arxiv.org/abs/1706.03762}
\showURL{%
\tempurl}


\bibitem[Xu et~al\mbox{.}(2025)]%
        {xu2025aesthetic}
\bibfield{author}{\bibinfo{person}{Junjie Xu}, \bibinfo{person}{Xingjiao Wu},
  \bibinfo{person}{Tanren Yao}, \bibinfo{person}{Zihao Zhang},
  \bibinfo{person}{Jiayang Bei}, \bibinfo{person}{Wu Wen}, {and}
  \bibinfo{person}{Liang He}.} \bibinfo{year}{2025}\natexlab{}.
\newblock \showarticletitle{Aesthetic Matters in Music Perception for Image
  Stylization: A Emotion-driven Music-to-Visual Manipulation}.
\newblock \bibinfo{journal}{\emph{arXiv preprint arXiv:2501.01700}}
  (\bibinfo{year}{2025}).
\newblock


\bibitem[Zhang et~al\mbox{.}(2024)]%
        {zhang2024minutes}
\bibfield{author}{\bibinfo{person}{Lintao Zhang}, \bibinfo{person}{Xiangcheng
  Du}, \bibinfo{person}{LeoWu TomyEnrique}, \bibinfo{person}{Yiqun Wang},
  \bibinfo{person}{Yingbin Zheng}, {and} \bibinfo{person}{Cheng Jin}.}
  \bibinfo{year}{2024}\natexlab{}.
\newblock \showarticletitle{Minutes to seconds: Speeded-up ddpm-based image
  inpainting with coarse-to-fine sampling}. In \bibinfo{booktitle}{\emph{2024
  IEEE International Conference on Multimedia and Expo (ICME)}}. IEEE,
  \bibinfo{pages}{1--6}.
\newblock


\bibitem[Zhou et~al\mbox{.}(2022)]%
        {zhou2022survey}
\bibfield{author}{\bibinfo{person}{Tianfei Zhou}, \bibinfo{person}{Fatih
  Porikli}, \bibinfo{person}{David~J Crandall}, \bibinfo{person}{Luc Van~Gool},
  {and} \bibinfo{person}{Wenguan Wang}.} \bibinfo{year}{2022}\natexlab{}.
\newblock \showarticletitle{A survey on deep learning technique for video
  segmentation}.
\newblock \bibinfo{journal}{\emph{IEEE transactions on pattern analysis and
  machine intelligence}} \bibinfo{volume}{45}, \bibinfo{number}{6}
  (\bibinfo{year}{2022}), \bibinfo{pages}{7099--7122}.
\newblock


\end{thebibliography}

\end{document}